\documentclass[11pt,a4paper]{article}
\usepackage[hyperref]{acl2021}
\usepackage{times}
\usepackage{latexsym}
\usepackage{bm}
\usepackage{amsmath,amssymb}
\usepackage{graphicx}
\usepackage{multirow}
\usepackage{subfigure}
\usepackage{color}
\usepackage{booktabs}

\usepackage{microtype}

\aclfinalcopy 
\def\aclpaperid{2656} 

\title{Crowdsourcing Learning as Domain Adaptation: {A} Case Study on Named Entity Recognition}

\author{
Xin Zhang$^1$, Guangwei Xu, Yueheng Sun$^2$, Meishan Zhang
$^{1}\thanks{~~Corresponding author.}$ , Pengjun Xie\\
$^1$School of New Media and Communication, Tianjin University, China\\
$^2$College of Intelligence and Computing, Tianjin University, China\\
\texttt{\{hsinz,yhs,zhangmeishan\}@tju.edu.cn}\\
\texttt{\{ahxgwOnePiece,xpjandy\}@gmail.com}\\
}

\date{}

\begin{document}
\maketitle
\begin{abstract}
Crowdsourcing is regarded as one prospective solution for effective supervised learning, aiming to build large-scale annotated training data by crowd workers.
Previous studies focus on reducing the influences from the noises of the crowdsourced annotations for supervised models.
We take a different point in this work, regarding all crowdsourced annotations as gold-standard with respect to the individual annotators.
In this way, we find that crowdsourcing could be highly similar to domain adaptation, and then the recent advances of cross-domain methods can be almost directly applied to crowdsourcing.
Here we take named entity recognition (NER) as a study case, suggesting an annotator-aware representation learning model that inspired by the domain adaptation methods which attempt to capture effective domain-aware features.
We investigate both unsupervised and supervised crowdsourcing learning, assuming that no or only small-scale expert annotations are available.
Experimental results on a benchmark crowdsourced NER dataset show that our method is highly effective, leading to a new state-of-the-art performance.
In addition, under the supervised setting, we can achieve impressive performance gains with only a very small scale of expert annotations.
\end{abstract}

\section{Introduction}
Crowdsourcing has gained a growing interest in the natural language processing (NLP) community, which helps hard NLP tasks such as named entity recognition \cite{finin-etal-2010-annotating,derczynski-etal-2016-broad}, part-of-speech tagging \cite{hovy-etal-2014-experiments}, relation extraction \cite{abad-etal-2017-self}, translation \cite{zaidan-callison-burch-2011-crowdsourcing}, argument retrieval \cite{dietrich2020fine}, and others \cite{snow-etal-2008-cheap,callison-burch-dredze-2010-creating} to collect a large scale dataset for supervised model training.
In contrast to the gold-standard annotations labeled by experts, the crowdsourced annotations can be constructed quickly at a low cost with masses of crowd annotators \cite{snow-etal-2008-cheap,nye-etal-2018-corpus}.
However, these annotations are relatively lower-quality with much-unexpected noise since the crowd annotators are not professional enough, which can make errors in complex and ambiguous contexts \cite{sheng2008get}.

Previous crowdsourcing learning models struggle to reduce the influences of noises of the crowdsourced annotations \cite{hsueh-etal-2009-data,raykar2012eli,hovy-etal-2013-learning,jamison-gurevych-2015-noise}.
Majority voting (MV) is one straightforward way to aggregate high-quality annotations, which has been widely adopted \cite{snow-etal-2008-cheap,fernandes2011learning,rodrigues2014sequence}, but it requires multiple annotations for a given input.
Recently, the majority of models concentrate on monitoring the distances between crowdsourced and gold-standard annotations, obtaining better performances than MV by considering the annotator information together \cite{nguyen-etal-2017-aggregating,simpson-gurevych-2019-bayesian,li-etal-2020-neural}.
Most of these studies assume the crowdsourced annotations as untrustworthy answers, proposing sophisticated strategies to recover the golden answers from crowdsourced labels.

\begin{figure} \centering
\includegraphics[scale=0.7]{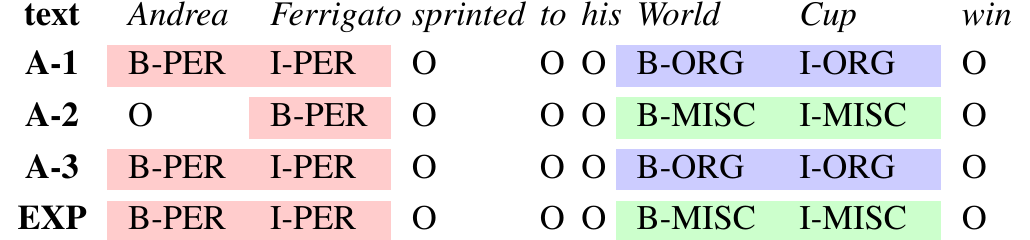}
\caption{A NER example with crowdsourced labels, A and EXP denote annotator and expert, respectively.}
\label{figure:example}
\end{figure}

In this work, we take a different view for crowdsourcing learning, regarding the crowdsourced annotations as the gold standard in terms of individual annotators.
In other words, we assume that all annotators (including experts) own their specialized understandings towards a specific task, and they annotate the task consistently according to their individual principles by the understandings, where the experts can reach an oracle principle by consensus.
The above view indicates that crowdsourcing learning aims to train a model based on the understandings of crowd annotators, and then test the model by the oracle understanding from experts.

Based on the assumption, we find that crowdsourcing learning is highly similar to domain adaptation, which is one important topic that has been investigated extensively for decades \cite{ben2006analysis,daume-iii-2007-frustratingly,chu-wang-2018-survey,jia-zhang-2020-multi}.
We treat each annotator as one domain specifically, and then crowdsourcing learning is essentially almost a multi-source domain adaptation problem.
Thus, one natural question arises: What is the performance when a state-of-the-art domain adaptation model is applied directly to crowdsourcing learning.

Here we take NER as a study case to investigate crowdsourcing learning as domain adaptation, considering that NER has been one popular task for crowdsourcing learning in the NLP community \cite{finin-etal-2010-annotating,rodrigues2014sequence,derczynski-etal-2016-broad}.
We suggest a state-of-the-art representation learning model that can effectively capture annotator(domain)-aware features.
Also, we investigate two settings of crowdsourcing learning, one being the unsupervised setting with no expert annotation, which has been widely studied before, and the other being the supervised setting where a certain scale of expert annotations exists, which is inspired by domain adaptation.

Finally, we conduct experiments on a benchmark crowdsourcing NER dataset \cite{tjong-kim-sang-de-meulder-2003-introduction,rodrigues2014sequence} to evaluate our methods.
We take a standard BiLSTM-CRF \cite{lample-etal-2016-neural} model with BERT \cite{devlin-etal-2019-bert} word representations as the baseline, and adapt it to our representation learning model.
Experimental results show that our method is able to model crowdsourced annotations effectively.
Under the unsupervised setting, our model can give a strong performance, outperforming previous work significantly.
In addition, the model performance can be greatly boosted by feeding with small-scale expert annotations, which can be a prospective direction for low-resource scenarios.

In summary, we make the following three major contributions:
\begin{itemize}
\setlength\itemsep{0em}
  \item[(1)] We present a different view of crowdsourcing learning, and propose to treat crowdsourcing learning as domain adaptation, which naturally connects the two important topics of machine learning for NLP.
  \item[(2)] We propose a novel method for crowdsourcing learning. Although the method is of a limited novelty for domain adaptation, it is the first work to crowdsourcing learning, and can achieve state-of-the-art performance on NER.
  \item[(3)] We introduce supervised crowdsourcing learning for the first time, which is borrowed from domain adaptation and would be a prospective solution for hard NLP tasks in practice.
\end{itemize}
We will release the code and detailed experimental settings at
\href{http://github.com/izhx/CLasDA}{github.com/izhx/CLasDA} under the Apache License 2.0 to facilitate future research.

\begin{figure} \centering
\includegraphics[scale=0.8]{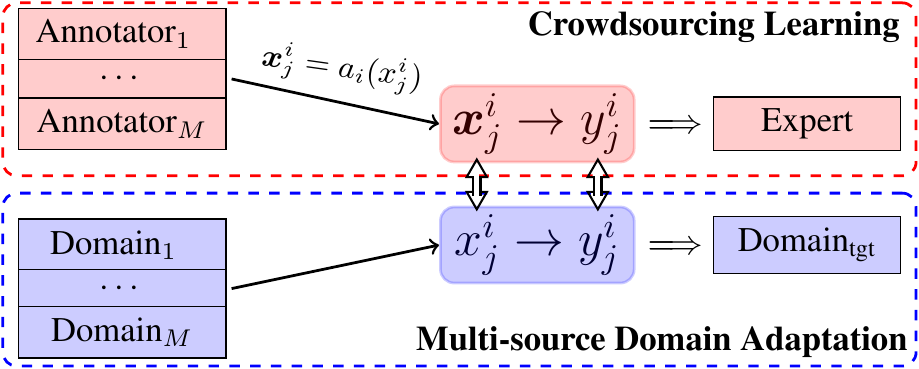}
\caption{Illustration of the connection between multi-source domain adaptation and crowdsourcing learning.}
\label{figure:cl2da}
\end{figure}

\section{The Basic Idea}
Here we describe the concepts of the domain adaptation and crowdsourcing learning in detail, and show how they are connected together.

\subsection{Domain Adaptation}
Domain adaptation happens when a supervised model trained on a fixed set of training corpus, including several specific domains, is required to test on a different domain
\cite{ben2006analysis,mansour2009domain}.
The scenario is quite frequent in practice, and thus has received extensive attention with massive investigations
\cite{csurka2017domain,ramponi-plank-2020-neural}.
The major problem lies in the different input distributions between source and target domains, leading to biased predictions over the inputs with a large gap to the source domains.

Here we focus on multi-source cross-domain adaptation, which would suit our next corresponding mostly.
Following \citet{mansour2009domain,zhao2019on}, the multi-source domain adaptation assumes a set of labeled examples from $M$ domains available, denoted by $D_{\text{src}} = \{ (X_i, Y_i) \}_{i=1}^M$,\footnote{
A domain is commonly defined as a distribution on the input data in many works, e.g., \citet{ben2006analysis}.
To make domain adaptation and crowdsourcing learning highly similar in formula, we follow \citet{zhao2019on}, defining a domain as a joint distribution on the input space $\mathcal{X}$ and the label space $\mathcal{Y}$.
Section \ref{domain-definition} gives a discussion of their connection.}
where $X_i = \{x_j^i\}_{j=1}^{N_i}$ and $Y_i = \{y_j^i\}_{j=1}^{N_i}$,\footnote{$N_{*}$ indicates the number of instances.}
and we aim to train a model on $D_{\text{src}}$ to adapt to a specific target domain with the help of a large scale raw corpus $X_{\text{tgt}} = \{ x_i \}_{i=1}^{N_t}$ of the target domain.

Note that under this setting, all $X$s, including source and target domains, are generated individually according to their unknown distributions, thus the abstract representations learned from the source domain dataset $D_{\text{src}}$ would inevitably be biased to the target domain, which is the primary reason for the degraded performance of the target domain \cite{huang-yates-2010-exploring,ganin2016domain}.
A number of domain adaptation models have struggled for better transferable high-level representations as domain shifts \cite{ramponi-plank-2020-neural}.

\subsection{Crowdsourcing Learning}
Crowdsourcing aims to produce a set of large-scale annotated examples created by crowd annotators, which is used to train supervised models for a given task \cite{raykar2010learning}.
As the majority of NLP models assume that gold-standard high-quality training corpora are already available \cite{manning1999foundations}, crowdsourcing learning has received much less interest than cross-domain adaptation, although the availability of these corpora is always not the truth.

Formally, under the crowdsourcing setting, we usually assume that there are a number of crowd annotators $A = \{ a_i \}_{i=1}^{M}$ (here we use the same $M$ as well as later superscripts in order to align with the domain adaptation), and all annotators should have a sufficient number of training examples by their different understandings for a given task, which are referred to as $D_{\text{crowd}} = \{ (X_i, Y_i) \}_{i=1}^M$ where $X_i = \{x_j^i\}_{j=1}^{N_i}$ and $Y_i = \{y_j^i\}_{j=1}^{N_i}$.
We aim to train a model on $D_{\text{crowd}}$ and adapt it to predict the expert outputs.
Note that all $X$s do not have significant differences in their distributions in this paradigm.

\paragraph{Crowdsourcing Learning as Domain adaptation}
By scrutinizing the above formalization, when we set all $X$s jointly with the annotators by using $\bm{x}_j^i = a_i(x_j^i)$, which indicates the contextualized understanding (a vectorial form is desirable here of the neural representations) of $x_j^i$ by the annotator $a_i$, then we would regard that $\bm{X}_i = \{a_i(x_j^i)\}_{j=1}^{N_i}$ is generated from different distributions as well.
In this way, we are able to connect crowdsourcing learning and domain adaptation together, as shown in Figure \ref{figure:cl2da}, based on the assumption that all $Y$s are gold-standard for crowdsourced annotations when crowd annotators are united as joint inputs.
And finally, we need to perform predictions by regarding $\bm{x}_{\text{expert}} = \text{expert}(x)$, and in particular, the learning of expert differs from that of the target domain in domain adaptation.

\section{A Case Study On NER}
In this section, we take NER as a case study,
which has been investigated most frequently in NLP \cite{yadav-bethard-2018-survey}, and propose a representation learning model mainly inspired by the domain adaptation model of
\cite{jia-etal-2019-cross} to perform crowdsourcing learning.
In addition, we introduce the unsupervised and supervised settings for crowdsourcing learning which are directly borrowed from the domain adaptation.

\begin{figure} \centering
\includegraphics[scale=1.1]{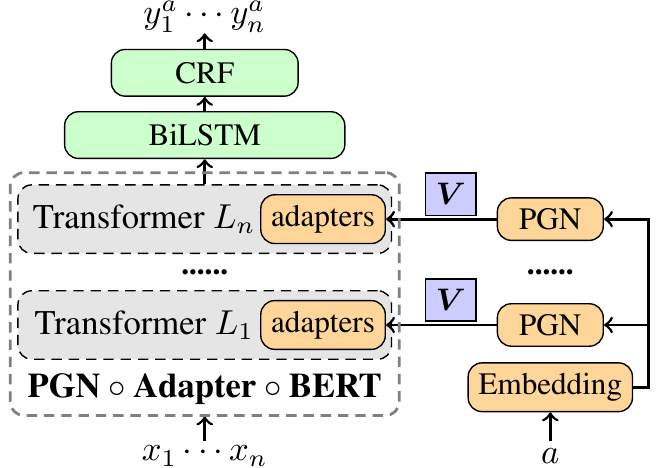}
\caption{The structure of our representation learning model, where the right orange part denotes the annotator switcher, and $\bm{V}$ denotes the generated adapter parameters by PGN. The transformer layers in gray are kept frozen in training, and other modules are trainable.}
\label{figure:model}
\end{figure}


\subsection{The Representation Learning Model}

We convert NER into a standard sequence labeling problem by using the BIO schema, following the majority of previous works, and extend a state-of-the-art BERT-BiLSTM-CRF model \cite{stephen2020robust} to our crowdsourcing learning.
Figure \ref{figure:model} shows the overall network structure of our representation learning model.
By using a sophisticated parameter generator module \cite{platanios-etal-2018-contextual}, it can capture annotator-aware features.
Following, we introduce the proposed model by four components: (1) word representation, (2) annotator switcher, (3) BiLSTM Encoding, and (4) CRF inference and training.

\paragraph{Word Representation}
Given a sentence of $n$ words $x = w_1\cdots w_n$, we first convert it to vectorial representations by BERT.
Different from the standard BERT exploration, here we use Adapter$\circ$BERT \cite{houlsby2019parameter}, where two extra adapter modules are inside each transformer layer.
The process can be simply formalized as:
\begin{equation}
    \bm{e}_1\cdots\bm{e}_n=\text{Adapter}\circ\text{BERT}(w_1\cdots w_n)
\end{equation}
where $\circ$ indicates an injection operation.
The detailed structure of the transformer with adapters is described in Appendix
\ref{appendix:transformer}.

Noticeably, the $\text{Adapter}\circ\text{BERT}$ method no longer needs fine-tuning the huge BERT parameters and can obtain comparable performance by adjusting the much lightweight adapter parameters instead.
Thus the representation can be more parameter efficient, and in this way we can easily extend the word representations to annotator-aware representations.

\paragraph{Annotator Switcher}
Our goal is to efficiently learn annotator-aware word representations, which can be regarded as contextualized understandings of individual annotators.
Hence, we introduce an annotator switcher to support $\text{Adapter}\circ\text{BERT}$ with annotator input as well, which is inspired by \citet{ustun-etal-2020-udapter}.
The key idea is to use Parameter Generation Network (PGN) \cite{platanios-etal-2018-contextual,jia-etal-2019-cross} to produce adapter parameters dynamically by input annotators.
In this way, our model can flexibly switch among different annotators.

Concretely, assuming that $\bm{V}$ is the vectorial form of all adapter parameters by a pack operation, which can also be unpacked to recover all adapter parameters as well, the PGN module is to generate $\bm{V}$ for $\text{Adapter}\circ\text{BERT}$ dynamically according the annotator inputs, as shown in Figure \ref{figure:model} by the right orange part.
The switcher can be formalized as:
\begin{equation}
\begin{split}
  \bm{x} & =\bm{r}'_1\cdots\bm{r}'_n \\
  & = \text{PGN}\circ\text{Adapter}\circ\text{BERT}(x, a) \\
              & = \text{Adapter}\circ\text{BERT} (x, \bm{V} = \bm{\Theta} \times \bm{e}^a),~~
\end{split}
\end{equation}
where $\bm{\Theta} \in \mathbb{R}^{|\bm{V}| \times |\bm{e}^a|}$ , $\bm{x} =\bm{r}'_1\cdots\bm{r}'_n$ is the annotator-aware representations of annotator $a$ for $x=w_1\cdots w_n$, and $\bm{e}^a$ is the annotator embedding.

\paragraph{BiLSTM Encoding}
$\text{Adapter}\circ\text{BERT}$ requires an additional task-oriented module for high-level feature extraction.
Here we exploit a single BiLSTM layer to achieve it: $\bm{h}_1\cdots \bm{h}_n = \text{BiLSTM}(\bm{x})$, which is used for next-step inference and training.

\paragraph{CRF Inference and Training}
We use CRF to calculate the score of a candidate sequential output $y=l_1\cdots l_n$ globally:
\begin{equation}\label{crf-score}
    \begin{split}
    & \bm{o}_i =\bm{W}^{\text{crf}}\bm{h}_i+\bm{b}^{\text{crf}} \\
    & \mathrm{score}(y|x,a) = \sum_{i=1}^{n}(\bm{T}[l_{i-1}, l_i]+\bm{o}_{i}[l_i]) \\
    \end{split}
\end{equation}
where $\bm{W}^{\text{crf}}$, $\bm{b}^{\text{crf}}$ and $\bm{T}$ are model parameters.

Given an input $(x, a)$, we perform inference by the Viterbi algorithm.
For training, we define a sentence-level cross-entropy objective:
\begin{equation}\label{crf_loss}
    \begin{split}
        & p(y^a|x,a) = \frac{\exp\big(\mathrm{score}(y^a|x,a)\big)}{\sum_{y}\exp\big(\mathrm{score}(y|x,a)\big)}  \\
        & \mathcal{L} = -\log p(y^a|x,a)
    \end{split}
\end{equation}
where $y^a$ is the gold-standard output of $x$ from $a$, $y$ belongs to all possible candidates, and $p(y^a|x,a)$ indicates the sentence-level probability.

\subsection{The Unsupervised Setting}
Here we introduce unsupervised crowdsourcing learning in alignment with unsupervised domain adaptation, assuming that no expert annotation is available, which is the widely-adopted setting of previous work of crowdsourcing learning \cite{sheng2008get,zhang2016learning,Sheng_Zhang_2019}.
This setting has a large divergence with domain adaptation in target learning.
In the unsupervised domain adaptation, the information of the target domain can be learned through a large-scale raw corpus \cite{ramponi-plank-2020-neural}, where there is no correspondence in the unsupervised crowdsourcing learning to learn information of experts.

To this end, here we suggest a simple and heuristic method to model experts by the specialty of crowdsourcing learning.
Intuitively, we expect that experts should approve the knowledge of the common consensus for a given task, and meanwhile, our model needs the embedding representation of experts for inference.
Thus, we can estimate the expert embedding by using the centroid point of all annotator embeddings:
\begin{equation}
    \bm{e}^{\text{expert}} = \frac{1}{|A|} \sum _{a \in A} \bm{e}^a
\end{equation}
where $A$ represents all annotators contributed to the training corpus.
This expert can be interpreted as the elected outcome by annotator voting with equal importance.
In this way, we perform the inference in unsupervised crowdsourcing learning by feeding $\bm{e}^{\text{expert}}$ as the annotator input.

\subsection{The Supervised Setting}
Inspired by the supervised domain adaptation, we also present the supervised crowdsourcing learning, which has been seldom concerned.
The setting is very simple, just by assuming that a certain scale of expert annotations is available.
In this way, we can learn the expert representation directly by supervised learning with our proposed model.

The supervised setting could be a more practicable scenario in real applications.
Intuitively, it should bring much better performance than the unsupervised setting with few shot expert annotations, which does not increase the overall annotation cost much.
In fact, during or after the crowdsourcing annotation process, we usually have a quality control module, which can help to produce silvery quality pseudo-expert annotations \cite{kittur2008crowd,lease2011on}.
Thus, the supervised setting can be highly valuable yet has been ignored mostly.

\section{Experiments}
\subsection{Setting}
\paragraph{Dataset}
We use the CoNLL-2003 NER English dataset \cite{tjong-kim-sang-de-meulder-2003-introduction} with crowdsourced annotations provided by \citet{rodrigues2018deep} to investigate our methods in both unsupervised and supervised settings.
The crowdsourced annotations consume 400 new articles, involving 5,985 sentences in practice, which are labeled by a total of 47 crowd annotators.
The total number of annotations is 16,878.
Thus the averaged number of annotated sentences per annotator is 359, which covers 6\% of the total sentences.
The dataset includes golden/expert annotations on the training sentences and a standard CoNLL-2003 test set for NER evaluation.

\paragraph{Evaluation}
The standard CoNLL-2003 evaluation metric is used to calculate the NER performance, reporting the entity-level precision (P), recall (R), and their F1 value.
All experiments of the same setting are conducted by five times, and the median outputs are used for performance reporting.
We exploit the pair-wise t-test for significance test, regarding two results significantly different when the p-value is below $10^{-5}$.

\setlength{\tabcolsep}{4.0pt}
\begin{table} \centering \small
\begin{tabular}{l|ccc}
\hline
Model & P & R & F1 \\
\hline \hline
 \multicolumn{4}{c}{Annotator-Agnostic} \\  \hline \hline
\texttt{ALL}              & 76.35 & \bf 72.47 & 74.36 \\
MV               & \bf 83.61 & 68.47 & \bf 75.28 \\
\hline \hline
\multicolumn{4}{c}{Annotator-Aware} \\  \hline \hline
\texttt{LC}               & 78.59 & 74.54 & 76.51 \\
\texttt{LC-cat}           & 74.34 & \bf 79.41 & 76.79 \\
\hline
\texttt{\bf This~Work} & \bf 78.84 & 75.67 & \bf 77.95 \\
\hline \hline
\multicolumn{4}{c}{Previous Work} \\ \hline \hline
\cite{rodrigues2014sequence} & 49.40 & 85.60 & 62.60 \\
LC \cite{nguyen-etal-2017-aggregating}     & \bf 82.38 & 62.10 & 70.82 \\
LC-cat \cite{nguyen-etal-2017-aggregating} & 79.61 & 62.87 & 70.26 \\
\cite{rodrigues2018deep} & 66.00 & 59.30 & 62.40 \\
\cite{simpson-gurevych-2019-bayesian}$^\dagger$ & 80.30 & \bf 74.80 & \bf 77.40 \\
\hline
\end{tabular}
\caption{The test results of the unsupervised setting, where the superscript
$^\dagger$ indicates that there exist differences in the test corpus.}
\label{table:unsupervised}
\end{table}

\begin{table*} \centering \small
\begin{tabular}{p{1.62cm}|ccc|ccc|ccc|ccc}
\hline
\multirow{2}{*}{Model} & \multicolumn{3}{c|}{1\%} & \multicolumn{3}{c|}{5\%} & \multicolumn{3}{c|}{25\%} & \multicolumn{3}{c}{100\%} \\ \cline{2-13}
& P & R & F1 & P & R & F1 & P & R & F1 & P & R & F1 \\
\hline \hline
\multicolumn{13}{c}{Annotator-Agnostic} \\  \hline \hline
\texttt{ALL}
& 75.08 & \bf 74.82 & \bf 74.95    & 76.18 & 75.71 & 75.94    & 78.64 & 78.93 & 78.78  & 86.65 & 82.29 & 84.42 \\ \cline{11-13}
\texttt{MV}
& \bf 83.87 & 67.37 & 74.72        & \bf 83.49 & 69.32 & 75.75    & \bf 84.77 &  79.43 &  82.01
& \multirow{2}{*}{\bf 89.28} & \multirow{2}{*}{\bf 89.77} & \multirow{2}{*}{\bf 89.52} \\ \cline{1-10}
\texttt{Gold}
& 69.52 & 75.41 & 72.35       & 76.70 & \bf 82.14 & \bf 79.33    & 81.32 & \bf 85.39 & \bf 83.31  &   &   &   \\
\hline \hline
\multicolumn{13}{c}{Annotator-Aware} \\  \hline \hline
\texttt{LC}
& 78.09 & 74.10 & 76.04   & 79.98 & 77.18 & 78.55  & 77.72 & 81.06 & 79.36  & 87.42 & 85.64 & 86.52 \\
\texttt{LC-cat}
& 75.37 & 78.54 & 76.92   & 74.24 & 81.32 & 77.62  & 76.88 & 81.37 & 78.96  & 88.25 & 86.03 & 87.13 \\
\hline
\texttt{\bf This~Work}
& \bf 80.06 & \bf 81.91 & \bf 80.97    & \bf 83.25 & \bf 85.36 & \bf 84.29  & \bf 85.19 & \bf 87.46 & \bf 86.31  & \bf 89.62 & \bf 90.51 & \bf 90.06 \\
\hline
\end{tabular}
\caption{The test results of the supervised setting, where we add different proportions of the most informative gold-standard (expert) annotations incrementally. Note that \texttt{MV} at 100\% is equivalent to the \texttt{gold} model, because all voted labels are substituted with gold-standard labels.}
\label{table:supervised}
\end{table*}

\paragraph{Baselines}
We re-implement several methods of previous work as baselines, and all the methods are based on Adapter$\circ$BERT-BiLSTM-CRF (no annotator switcher inside) for fair comparisons.

For both the unsupervised and supervised settings, we consider the following baseline models:
\begin{itemize}
\setlength\itemsep{0em}
    \item \texttt{ALL}: which treats all annotations equally, ignoring the annotator information no matter crowd or expert.
    \item \texttt{MV}: which is borrowed from \newcite{rodrigues2014sequence}, where aggregated labels are produced by token level majority voting. In particular,
    the gold-standard labels are used instead if they are available for a specific sentence during the supervised crowdsourcing learning.
    \item \texttt{LC}: which is proposed by \newcite{nguyen-etal-2017-aggregating}, where the annotator bias to the gold-standard labels is explicitly modeled at the CRF layer for each crowd annotator, and specifically, the expert is with zero bias.
    \item \texttt{LC-cat}: which is also presented by \newcite{nguyen-etal-2017-aggregating} as a baseline to \texttt{LC}, where the annotator bias is modeled at the BiLSTM layer instead
    and also the expert bias is set to zero.\footnote{Note that although \texttt{LC-cat} is not as expected as \texttt{LC} in \cite{nguyen-etal-2017-aggregating}, our results show that \texttt{LC-cat} is slightly better based on Adapter$\circ$BERT-BiLSTM-CRF.}
\end{itemize}
Notice that \texttt{ALL} and \texttt{MV} are annotator-agnostic models, which exploit no information specific to the individual annotators, while the other three models are all annotator-aware models, where the annotator information is used by different ways.

\paragraph{Hyper-parameters}
We offer all detailed settings of Hyper-parameters in Appendix \ref{appendix:hyper-param}.

\subsection{Unsupervised Results}
Table \ref{table:unsupervised} shows the test results of the unsupervised setting.
As a whole, we can see that our representation learning model (i.e., \texttt{This Work}) borrowed from domain adaptation can achieve the best performance, resulting in an F1 score of $77.95$, significantly better than the second-best model \texttt{LC-cat} (i.e., $77.95-76.79=1.16$).
The result indicates the advantage of our method over the other models.

By examining the results in-depth, we can find that the annotator-aware model is significantly better than the annotator-agnostic models, demonstrating that the annotator information is highly helpful for crowdsourcing learning.
The observation further shows the reasonableness by aligning annotators to domains, since domain information is also useful for domain adaptation.
In addition, the better performance of our representation learning method among the annotator-aware models indicates that our model can capture annotator-aware information more effectively because our start point is totally different.
We do not attempt to model the expert labels based on crowdsourcing annotations.

Further, we observe that several models show better precision values, while others give better recall values.
A high precision but low recall indicates that the model is conservative in detecting named entities, and vice the reverse.
Our proposed model is able to balance the two directions better, with the least gap between them.
Also, the results imply that there is still much space for future development, and the recent advances of domain adaptation might offer good avenues.

Finally, we compare our results with previous studies.
As shown, our model can obtain the best performance in the literature.
In particular, by comparing our results with the original performances reported in
\citet{nguyen-etal-2017-aggregating},
we can see that our re-implementation is much better than theirs.
The major difference lies in the exploration of BERT in our model, which brings improvements closed to 6\% for both \texttt{LC} and \texttt{LC-cat}.

\subsection{Supervised Results}
To investigate the supervised setting, we assume that expert annotations (ground truths) of all crowdsourcing sentences are available.
Besides exploring the full expert annotations, we study another three different scenarios by incrementally adding the expert annotations into the unsupervised setting, aiming to study the effectiveness of our model with small expert annotations as well.
Concretely, we assume proportions of 1\%, 5\%, 25\%, and 100\% of the expert annotations available.\footnote{Intuitively, if expert annotations are involved, we should intentionally choose the more informative inputs for annotations, which can reduce the overall cost to meet a certain performance standard. Thus, we can fully demonstrate the effectiveness of crowdsourced annotations under the semi-supervised setting. Here we try to choose the most informative labeled instances for the 1\%, 5\%, and 25\% settings.}
Table \ref{table:supervised} shows all the results, including our four baselines and an \texttt{gold} model based on only expert annotations for comparisons.
Overall, we can see that our representation learning model can bring the best performances for all scenarios, demonstrating its effectiveness in the supervised learning as well.

Next, by comparing annotator-agnostic and annotator-aware models, we can see that annotator-aware models are better, which is consistent with the unsupervised setting.
More interestingly, the results show that \texttt{All} is better than \texttt{gold} with very small-scale expert annotations (1\% and 5\%), and the tendency is reversed only when there are sufficient expert annotations (25\% and 100\%).
The observation indicates that crowdsourced annotations are always helpful when golden annotations are not enough.
In addition, it is easy to understand that \texttt{MV} is worse than \texttt{gold} since the latter has a higher-quality of the training corpus.

Further, we can find that even the annotator-aware \texttt{LC} and \texttt{LC-cat} models are unable to obtain any positive influence compared with \texttt{gold}, which demonstrates that distilling ground-truths from the crowdsourcing annotations might not be the most promising solution.
While our representation learning model can give consistently better results than \texttt{gold}, indicating that crowdsourced annotations are always helpful by our method.
By regarding crowdsourcing learning as domain adaptation, we no longer take crowdsourced annotations as noise, and on the contrary, they are treated as transferable knowledge, similar to the relationship between the source domains and the target domain.
Thus they could always be useful in this way.


\subsection{Analysis}
To better understand our idea and model in-depth, we conducted the following fine-grained analyses.\footnote{
In addition, we could not perform the ablation study of our model because it is not an incremental work.}

\begin{figure}
\centering
\subfigure[0\%]{\label{figure:vis-un}
\includegraphics[scale=0.4]{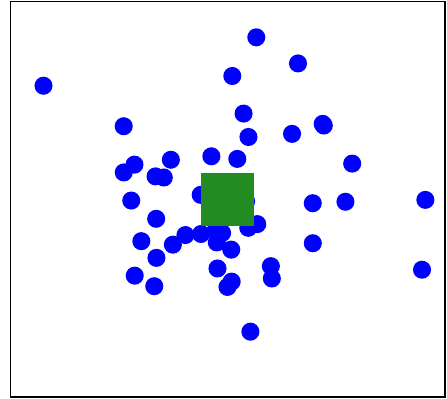}
}
\hspace{-0.3cm}
\subfigure[5\%]{\label{figure:vis-s05}
\includegraphics[scale=0.4]{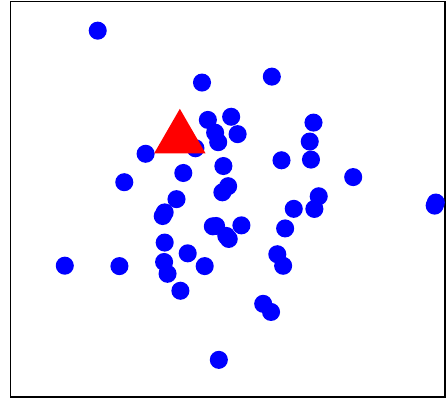}
}
\hspace{-0.3cm}
\subfigure[25\%]{\label{figure:vis-s25}
\includegraphics[scale=0.4]{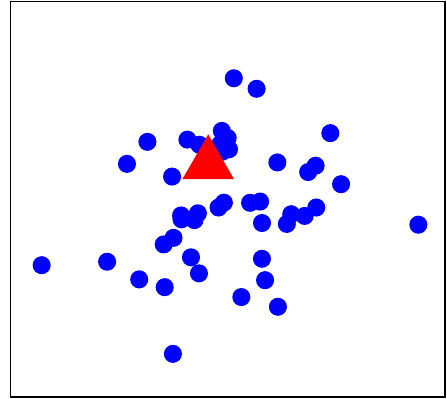}
}
\hspace{-0.3cm}
\subfigure[100\%]{\label{figure:vis-s}
\includegraphics[scale=0.4]{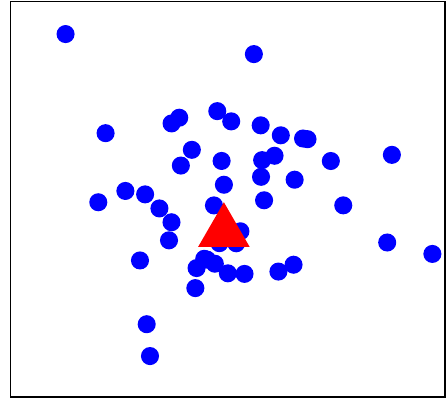}
}
\caption{The visualization of annotator embeddings by dimensionality reduction with PCA. Out designed unsupervised (0\%) expert is consistent with the well-learned one (100\%). With the expert annotations increases, the learned expert becomes more accurate.}
\label{figure:vis}
\end{figure}

\paragraph{Visualization of Annotator Embeddings}
Our representation learning model is able to learn annotator embeddings through the task objective.
It is interesting to visualize these embeddings to check their distributions, which can reflect the relationships between the individual annotators.
Figure \ref{figure:vis} shows the visualization results after Principal Component Analysis (PCA) dimensionality reduction, where the unsupervised and three supervised scenarios are investigated.\footnote{
The 1\% setting is excluded for its incapability to capture the relationship between the expert and crowd annotators with such small expert annotations.}
As shown, we can see that most crowd annotators are distributed in a concentrated area for all scenarios, indicating that they are able to share certain common characteristics of task understanding.

Further, we focus on the relationship between expert and crowd annotators, and the results show two interesting findings.
First, the heuristic expert of our unsupervised learning is almost consistent with that of the supervised learning of the whole expert annotations (100\%), which indicates that our unsupervised expert estimation is perfectly good.
Second, the visualization shows that the relationship between expert and crowd annotators could be biased when expert annotations are not enough.
As the size of expert annotations increases, their connection might be more accurate gradually.

\setlength{\tabcolsep}{8.0pt}
\begin{table} \centering \small
\begin{tabular}{l|ccc|c}
\hline
Model & P & R & F1 & Gold(5\%)\\
\hline \hline
\texttt{ALL}              & 67.02 & 69.31 & 68.15 & \multirow{5}{*}{\bf 79.33}\\
MV               & 72.24 & 69.49 & 70.88 &\\
\texttt{LC}               & 72.34 & 70.48 & 71.35 &\\
\texttt{LC-cat}           & \bf 72.76 & \bf 71.78 & \bf 72.26 &\\
\cline{1-4}
\texttt{\bf This~Work} & \bf 80.78 & \bf 73.78 & \bf 77.12 & \\
\hline
\end{tabular}
\caption{The performance of training on 85\% and testing on 15\% of the crowdsourced annotations.}
\label{table:annotator}
\end{table}

\paragraph{The Predictability of Crowdsourcing Annotations}
Our primary assumption is based on that all crowdsourced annotations are regarded as the gold-standard with respect to the crowd annotators, which naturally indicates that these annotations are predictable.
Here we conduct analysis to verify the assumption by a new task to predicate the crowdsourced annotations,
Concretely, we divide the annotations into two sections, where 85\% of them are used as the training and the remaining are used for testing, and then we apply our baseline and proposed models to learn and evaluate.

Table \ref{table:annotator} shows the results.
As shown, our model can achieve the best performance by an F1 score of 77.12\%, and the other models are significantly worse (at least $4.86$ drops by F1).
Considering that the proportion of the averaged training examples per annotator over the full 5,985 sentences is only 5\%,\footnote{
The value can be directly calculated ($0.06*0.85\approx0.05$).}
we exploit the \texttt{gold} model of the 5\% expert annotations for reference.
We can see that the gap between them is small (77.12\% v.s. 79.33\%), which indicates that our assumption is acceptable as a whole.
The other models could be unsuitable for our assumption due to the poor performance induced by their modeling strategies.

\begin{figure}
\centering
\subfigure[0\%]{\label{figure:db-un}
\includegraphics[scale=0.57]{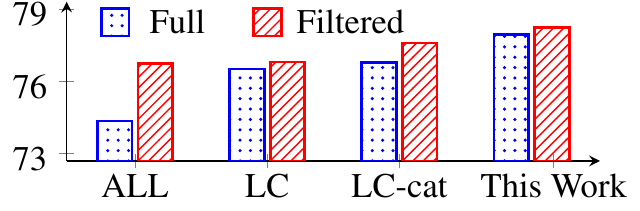}
}
\hspace{-0.3cm}
\subfigure[1\%]{\label{figure:db-s05}
\includegraphics[scale=0.57]{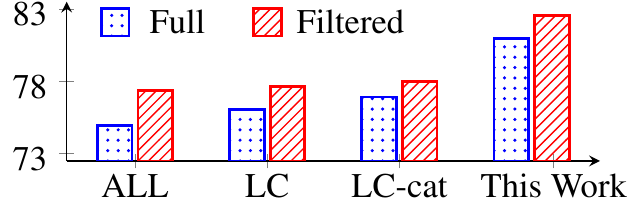}
}
\subfigure[5\%]{\label{figure:db-s25}
\includegraphics[scale=0.57]{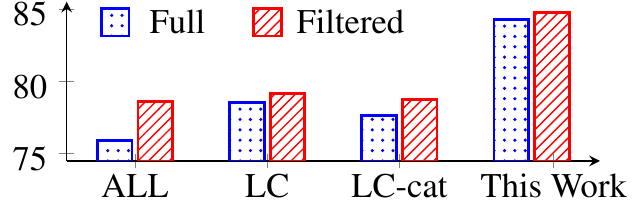}
}
\hspace{-0.3cm}
\subfigure[25\%]{\label{figure:db-s}
\includegraphics[scale=0.57]{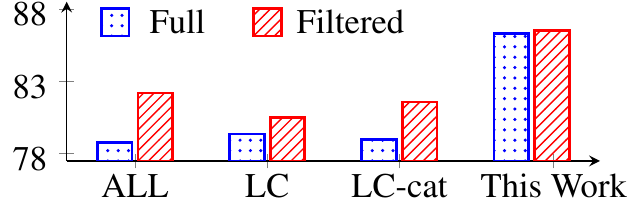}
}
\caption{Comparisons by F1 scores between full and filtered crowdsourced annotations (i.e., excluding unreliable annotators). We compute F1 values of each annotator with respect to the gold-standard labels, and filter out 10 annotators with lowest scores.}
\label{figure:exclude}
\end{figure}

\paragraph{The Impact of Unreliable Annotators}
Handling unreliable annotators, such as spammers, is a practical and common issue in Crowdsourcing \cite{vikas2012eliminating}.
Obviously, regarding crowd annotations as untrustworthy answers is more considerate to this problem.
In contrast, our assumption might be challenged because these unreliable annotators are discrepant in their own annotations.
To show the influence of unreliable annotators, we filter out several unreliable annotators in the corpus, and reevaluate the performance for the low-resource supervised and unsupervised scenarios on the remaining annotations.

Figure \ref{figure:exclude} shows the comparison results of the original corpus and the filtered corpus.\footnote{
\texttt{MV} is not included because a proportion of instances are unable to obtain aggregated answers.}
First, we can find that improved performance can be achieved in all cases, indicating excluding these unreliable annotations is helpful for crowdsourcing.
Second, the \texttt{LC} and \texttt{LC-cat} model give smaller score differences compared with the \texttt{ALL} model between these two kinds of results, which verified that they are considerate to unreliable annotators.
Third, our model also performs robustly, it can cope with this practical issue in a certain degree as well.

\setlength{\tabcolsep}{4.0pt}
\begin{table} \centering \small
\begin{tabular}{l|cccc}
\hline
Data & Full & Excluded & Part-1 & Part-2 \\
\hline
Model & \multicolumn{4}{c}{F1}  \\
\hline \hline
\texttt{ALL}              & 74.36 & 76.73 & 74.66 & 75.92 \\
\hline
\texttt{LC}               & 76.51 & 76.80 & 75.29 & 76.70 \\
\texttt{LC-cat}           & 76.79 & 77.59 & 74.86 & 76.02 \\
\hline
\texttt{\bf This~Work}    & 77.95 & 78.23 & 77.41 & 77.58 \\
\hline
\end{tabular}
\caption{The unsupervised test results of differently sampled datasets.
The Full is original results in Table \ref{table:unsupervised}.
The Excluded is the filtered corpus in Figure \ref{figure:exclude}.
The Part-1 and Part-2 are both consist of $13$ annotators.
Part-1 have $1800$ texts with $6275$ crowd annotations, each text is labeled by at least $3$ annotators.
These numbers of Part-2 are $2192$, $5582$, and $2$, respectively.}
\label{table:sampled}
\end{table}

\paragraph{Results on The Sampled Annotators and Annotations}
The above analysis shows the benefit of removing unreliable annotators, which reduces a small number of annotators and annotations.
A problem arises naturally: will the performance be consistent if we sample a small proportion of annotators?
To verify it, we sampled two sub-set from the crowdsourced training corpus and re-train our model as well as baselines.
Table \ref{table:sampled} shows the evaluation results of re-trained models on the standard test set in unsupervised setting.
We also add our main result for the comparison.
As shown, all sampled datasets demonstrate similar trends with the main result (denoted as \texttt{Full}).
The supervised results are consistent with our main result as well, which are not listed due to space reasons.

\subsection{The Discussion of Domain Definitions}
\label{domain-definition}
The most widely used definition of a domain is the distribution on the input space $\mathcal{X}$.
\citet{zhao2019on} define a domain $D$ as the pair of a distribution $\mathcal{D}$ on the input space $\mathcal{X}$ and a labeling function $f:\mathcal{X} \to \mathcal{Y}$, i.e., domain $D = \langle \mathcal{D}, f \rangle$.

In this work, we assume each annotator is a unique labeling function $a:\mathcal{X} \to \mathcal{Y}$.
Uniting each annotator and the instances he/she labeled, we can result in a number of domains $\{\langle \mathcal{D}_i, a_i \rangle\}_{i=1}^{|A|}$, where $A$ represents all annotators.
Then the crowdsourcing learning can be interpreted by the later definition, i.e., learning from these crowd annotators/domains and predicting the labels of raw inputs (sampled from the raw data distribution $\mathcal{D}_\text{expert}$) in expert annotator/domain $\langle \mathcal{D}_\text{expert}, \text{expert} \rangle$.
To unify the definition in a single distribution, we directly define a domain as the joint distribution on the input space $\mathcal{X}$ and the label space $\mathcal{Y}$.

In addition, we can align to the former definition by using the representation outputs $\bm{x}^i = a_i(x)$ as the data input, which shows different distributions for the same sentence towards different annotators.
Thus, each source domain $D_i$ is the distibution of $\bm{x}^i$, and we need learn the expert representations $\bm{x}^\text{expert}$ to perform inference on the unlabled texts.

\section{Related Work}
\subsection{Crowdsourcing Learning}
Crowdsourcing is a cheap and popular way to collect large-scale labeled data, which can facilitate the model training for hard tasks that require supervised learning \cite{wang2016cost,Sheng_Zhang_2019}.
In particular, crowdsourced data is often regarded as low-quality, including much noise regarding expert annotations as the gold-standard.
Initial studies of crowdsourcing learning try to arrive at a high-quality corpus by majority voting or control the quality by sophisticated strategies during the crowd annotation process \cite{khattak2011quality,liu2017improving,tang2011semi}.

Recently, the majority work focuses on full exploration of all annotated corpus by machine learning models, taking the information from crowd annotators into account including annotator reliability \cite{rodrigues2014sequence}, annotator accuracy \cite{huang-etal-2015-estimation},  worker-label confusion matrix \cite{nguyen-etal-2017-aggregating}, and sequential confusion matrix \cite{simpson-gurevych-2019-bayesian}.

In this work, we present a totally different viewpoint for crowdsourcing, regarding all crowdsourced annotations as golden in terms of individual annotators, just like the primitive gold-standard labels corresponded to the experts, and further propose a domain adaptation paradigm for crowdsourcing learning.

\subsection{Domain Adaptation}
Domain adaptation has been studied extensively to reduce the performance gap between the resource-rich and resource-scarce domains \cite{ben2006analysis,mansour2009domain}, which has also received great attention in the NLP community \cite{daume-iii-2007-frustratingly,jiang-zhai-2007-instance,finkel-manning-2009-hierarchical,glorot2011domain,chu-wang-2018-survey,ramponi-plank-2020-neural}.
Typical methods include self-training to produce pseudo training instances for the target domain \cite{yu-etal-2015-domain} and representation learning to capture transferable features across the source and target domains \cite{sener2016learning}.

In this work, we make correlations between domain adaptation and crowdsourcing learning, enabling crowdsourcing learning to benefit from the advances of domain adaptation, and then present a representation learning model borrowed from \citet{jia-etal-2019-cross} and \citet{ustun-etal-2020-udapter}.

\subsection{Named Entity Recognition}
NER is a fundamental and challenging task of NLP \cite{yadav-bethard-2018-survey}.
The BiLSTM-CRF \cite{lample-etal-2016-neural} architecture, as well as BERT \cite{devlin-etal-2019-bert}, are able to bring state-of-the-art performance in the literature \cite{jia-etal-2019-cross,wang-etal-2020-multi-domain,jia-zhang-2020-multi}.
\citet{stephen2020robust} exploits the BERT-BiLSTM-CRF model, achieving strong performance on NER.

In addition, NER has been widely adopted as crowdsourcing learning as well \cite{finin-etal-2010-annotating,rodrigues2014sequence,derczynski-etal-2016-broad,yang2018adversarial}.
Thus, we exploit NER as a case study following these works, and take a BERT-BiLSTM-CRF model as the basic model for our annotator-aware extension.

\section{Conclusion and Future Work}
We studied the connection between crowdsourcing learning and domain adaptation, and then proposed to treat crowdsourcing learning as a domain adaptation problem.
Following, we took NER as a case study, suggesting a representation learning model from recent advances of domain adaptation for crowdsourcing learning.
By this case study, we introduced unsupervised and supervised crowdsourcing learning, where the former is a widely-studied setting while the latter has been seldom investigated.
Finally, we conducted experiments on a widely-adopted benchmark dataset for crowdsourcing NER, and the results show that our representation learning model is highly effective in unsupervised learning, achieving the best performance in the literature.
In addition, the supervised learning with a very small scale of expert annotations can boost the performance significantly.

Our work sheds light on the application of effective domain adaptation models on crowdsourcing learning.
There are still many other sophisticated cross-domain models, such as adversarial learning \cite{ganin2016domain} and self-training \cite{yu-etal-2015-domain}.
Future work may include how to apply these advances to crowdsourcing learning properly.

\section*{Acknowledgments}
We thank all reviewers for their hard work.
This research is supported by grants from the National Key Research and Development Program of China (No. 2018YFC0832101)
and the fonds of Beijing Advanced Innovation Center for Language Resources under Grant TYZ19005.

\section*{Ethical Impact}
We present a different view of crowdsourcing learning and propose to treat it as domain adaptation, showing the connection between these two topics of machine learning for NLP.
In this view, many sophisticated cross-domain models could be applied to crowdsourcing learning.
Moreover, the motivation that regarding all crowdsourced annotations as gold-standard to the corresponding annotators, also sheds light on introducing other transfer learning techniques in future work.

The above idea and our proposed representation learning model for crowdsourcing sequence labeling, are totally agnostic to any private information of annotators.
And we do not use any sensitive information, bu only the ID of annotators, in problem modeling and learning.
The crowdsourced CoNLL English NER data also anonymized annotators.
There will be no privacy issues in the future.

\bibliographystyle{acl_natbib}
\bibliography{acl2021}

\begin{thebibliography}{59}
\expandafter\ifx\csname natexlab\endcsname\relax\def\natexlab#1{#1}\fi

\bibitem[{Abad et~al.(2017)Abad, Nabi, and Moschitti}]{abad-etal-2017-self}
Azad Abad, Moin Nabi, and Alessandro Moschitti. 2017.
\newblock \href {https://doi.org/10.18653/v1/P17-2082} {Self-crowdsourcing
  training for relation extraction}.
\newblock In \emph{Proceedings of the ACL: Short Papers}.

\bibitem[{Ben{-}David et~al.(2006)Ben{-}David, Blitzer, Crammer, and
  Pereira}]{ben2006analysis}
Shai Ben{-}David, John Blitzer, Koby Crammer, and Fernando Pereira. 2006.
\newblock \href
  {https://proceedings.neurips.cc/paper/2006/hash/b1b0432ceafb0ce714426e9114852ac7-Abstract.html}
  {Analysis of representations for domain adaptation}.
\newblock In \emph{Proceedings of the Twentieth Annual Conference on Neural
  Information Processing Systems}, pages 137--144. {MIT} Press.

\bibitem[{Callison-Burch and
  Dredze(2010)}]{callison-burch-dredze-2010-creating}
Chris Callison-Burch and Mark Dredze. 2010.
\newblock \href {https://www.aclweb.org/anthology/W10-0701} {Creating speech
  and language data with {A}mazon{'}s {M}echanical {T}urk}.
\newblock In \emph{Proceedings of the NAACL-HLT 2010 Workshop on Creating
  Speech and Language Data with {A}mazon{'}s Mechanical Turk}, pages 1--12.

\bibitem[{Chu and Wang(2018)}]{chu-wang-2018-survey}
Chenhui Chu and Rui Wang. 2018.
\newblock \href {https://www.aclweb.org/anthology/C18-1111} {A survey of domain
  adaptation for neural machine translation}.
\newblock In \emph{Proceedings of COLING}, pages 1304--1319.

\bibitem[{Csurka(2017)}]{csurka2017domain}
Gabriela Csurka. 2017.
\newblock Domain adaptation for visual applications: A comprehensive survey.
\newblock \emph{arXiv preprint arXiv:1702.05374}.

\bibitem[{Daum{\'e}~III(2007)}]{daume-iii-2007-frustratingly}
Hal Daum{\'e}~III. 2007.
\newblock \href {https://www.aclweb.org/anthology/P07-1033} {Frustratingly easy
  domain adaptation}.
\newblock In \emph{Proceedings of ACL}, pages 256--263.

\bibitem[{Derczynski et~al.(2016)Derczynski, Bontcheva, and
  Roberts}]{derczynski-etal-2016-broad}
Leon Derczynski, Kalina Bontcheva, and Ian Roberts. 2016.
\newblock \href {https://www.aclweb.org/anthology/C16-1111} {Broad {T}witter
  corpus: A diverse named entity recognition resource}.
\newblock In \emph{Proceedings of the COLING: Technical Papers}, pages
  1169--1179.

\bibitem[{Devlin et~al.(2019)Devlin, Chang, Lee, and
  Toutanova}]{devlin-etal-2019-bert}
Jacob Devlin, Ming-Wei Chang, Kenton Lee, and Kristina Toutanova. 2019.
\newblock \href {https://doi.org/10.18653/v1/N19-1423} {{BERT}: Pre-training of
  deep bidirectional transformers for language understanding}.
\newblock In \emph{Proceedings of NAACL-HLT}.

\bibitem[{Fernandes and Brefeld(2011)}]{fernandes2011learning}
Eraldo~R. Fernandes and Ulf Brefeld. 2011.
\newblock \href {https://doi.org/10.1007/978-3-642-23780-5\_36} {Learning from
  partially annotated sequences}.
\newblock In \emph{ECML-PKDD}, volume 6911 of \emph{Lecture Notes in Computer
  Science}, pages 407--422. Springer.

\bibitem[{Finin et~al.(2010)Finin, Murnane, Karandikar, Keller, Martineau, and
  Dredze}]{finin-etal-2010-annotating}
Tim Finin, William Murnane, Anand Karandikar, Nicholas Keller, Justin
  Martineau, and Mark Dredze. 2010.
\newblock \href {https://www.aclweb.org/anthology/W10-0713} {Annotating named
  entities in {T}witter data with crowdsourcing}.
\newblock In \emph{Proceedings of the NAACL-HLT 2010 Workshop on Creating
  Speech and Language Data with {A}mazon{'}s Mechanical Turk}.

\bibitem[{Finkel and Manning(2009)}]{finkel-manning-2009-hierarchical}
Jenny~Rose Finkel and Christopher~D. Manning. 2009.
\newblock \href {https://www.aclweb.org/anthology/N09-1068} {Hierarchical
  {B}ayesian domain adaptation}.
\newblock In \emph{Proceedings of HLT-NAACL}, pages 602--610.

\bibitem[{Ganin et~al.(2016)Ganin, Ustinova, Ajakan, Germain, Larochelle,
  Laviolette, Marchand, and Lempitsky}]{ganin2016domain}
Yaroslav Ganin, Evgeniya Ustinova, Hana Ajakan, Pascal Germain, Hugo
  Larochelle, Fran{\c{c}}ois Laviolette, Mario Marchand, and Victor~S.
  Lempitsky. 2016.
\newblock \href {http://jmlr.org/papers/v17/15-239.html} {Domain-adversarial
  training of neural networks}.
\newblock \emph{J. Mach. Learn. Res.}, 17:59:1--59:35.

\bibitem[{Glorot et~al.(2011)Glorot, Bordes, and Bengio}]{glorot2011domain}
Xavier Glorot, Antoine Bordes, and Yoshua Bengio. 2011.
\newblock \href {https://icml.cc/2011/papers/342\_icmlpaper.pdf} {Domain
  adaptation for large-scale sentiment classification: {A} deep learning
  approach}.
\newblock In \emph{Proceedings of ICML}, pages 513--520.

\bibitem[{Houlsby et~al.(2019)Houlsby, Giurgiu, Jastrzebski, Morrone,
  De~Laroussilhe, Gesmundo, Attariyan, and Gelly}]{houlsby2019parameter}
Neil Houlsby, Andrei Giurgiu, Stanislaw Jastrzebski, Bruna Morrone, Quentin
  De~Laroussilhe, Andrea Gesmundo, Mona Attariyan, and Sylvain Gelly. 2019.
\newblock \href {http://proceedings.mlr.press/v97/houlsby19a.html}
  {Parameter-efficient transfer learning for {NLP}}.
\newblock In \emph{Proceedings of ICML}, pages 2790--2799.

\bibitem[{Hovy et~al.(2013)Hovy, Berg-Kirkpatrick, Vaswani, and
  Hovy}]{hovy-etal-2013-learning}
Dirk Hovy, Taylor Berg-Kirkpatrick, Ashish Vaswani, and Eduard Hovy. 2013.
\newblock \href {https://www.aclweb.org/anthology/N13-1132} {Learning whom to
  trust with {MACE}}.
\newblock In \emph{Proceedings of the NAACL-HLT}.

\bibitem[{Hovy et~al.(2014)Hovy, Plank, and
  S{\o}gaard}]{hovy-etal-2014-experiments}
Dirk Hovy, Barbara Plank, and Anders S{\o}gaard. 2014.
\newblock \href {https://doi.org/10.3115/v1/P14-2062} {Experiments with
  crowdsourced re-annotation of a {POS} tagging data set}.
\newblock In \emph{Proceedings of ACL}.

\bibitem[{Hsueh et~al.(2009)Hsueh, Melville, and
  Sindhwani}]{hsueh-etal-2009-data}
Pei-Yun Hsueh, Prem Melville, and Vikas Sindhwani. 2009.
\newblock \href {https://www.aclweb.org/anthology/W09-1904} {Data quality from
  crowdsourcing: A study of annotation selection criteria}.
\newblock In \emph{Proceedings of the {NAACL} {HLT} 2009 Workshop on Active
  Learning for Natural Language Processing}, pages 27--35.

\bibitem[{Huang and Yates(2010)}]{huang-yates-2010-exploring}
Fei Huang and Alexander Yates. 2010.
\newblock \href {https://www.aclweb.org/anthology/W10-2604} {Exploring
  representation-learning approaches to domain adaptation}.
\newblock In \emph{Proceedings of the 2010 Workshop on Domain Adaptation for
  Natural Language Processing}.

\bibitem[{Huang et~al.(2015)Huang, Zhong, and
  Passonneau}]{huang-etal-2015-estimation}
Ziheng Huang, Jialu Zhong, and Rebecca~J. Passonneau. 2015.
\newblock \href {https://doi.org/10.18653/v1/D15-1261} {Estimation of discourse
  segmentation labels from crowd data}.
\newblock In \emph{Proceedings of the EMNLP}, pages 2190--2200.

\bibitem[{Jamison and Gurevych(2015)}]{jamison-gurevych-2015-noise}
Emily Jamison and Iryna Gurevych. 2015.
\newblock \href {https://doi.org/10.18653/v1/D15-1035} {Noise or additional
  information? leveraging crowdsource annotation item agreement for natural
  language tasks.}
\newblock In \emph{Proceedings of the EMNLP}, pages 291--297.

\bibitem[{Jia et~al.(2019)Jia, Liang, and Zhang}]{jia-etal-2019-cross}
Chen Jia, Xiaobo Liang, and Yue Zhang. 2019.
\newblock \href {https://doi.org/10.18653/v1/P19-1236} {Cross-domain {NER}
  using cross-domain language modeling}.
\newblock In \emph{Proceedings of ACL}, pages 2464--2474.

\bibitem[{Jia and Zhang(2020)}]{jia-zhang-2020-multi}
Chen Jia and Yue Zhang. 2020.
\newblock \href {https://doi.org/10.18653/v1/2020.acl-main.524} {Multi-cell
  compositional {LSTM} for {NER} domain adaptation}.
\newblock In \emph{Proceedings of ACL}, pages 5906--5917.

\bibitem[{Jiang and Zhai(2007)}]{jiang-zhai-2007-instance}
Jing Jiang and ChengXiang Zhai. 2007.
\newblock \href {https://www.aclweb.org/anthology/P07-1034} {Instance weighting
  for domain adaptation in {NLP}}.
\newblock In \emph{Proceedings of ACL}, pages 264--271.

\bibitem[{Khattak and Salleb-Aouissi(2011)}]{khattak2011quality}
Faiza~Khan Khattak and Ansaf Salleb-Aouissi. 2011.
\newblock Quality control of crowd labeling through expert evaluation.
\newblock In \emph{Proceedings of the NIPS 2nd Workshop on Computational Social
  Science and the Wisdom of Crowds}, volume~2, page~5.

\bibitem[{Kittur et~al.(2008)Kittur, Chi, and Suh}]{kittur2008crowd}
Aniket Kittur, Ed~H. Chi, and Bongwon Suh. 2008.
\newblock \href {https://doi.org/10.1145/1357054.1357127} {Crowdsourcing user
  studies with mechanical turk}.
\newblock In \emph{Proceedings of the 2008 Conference on Human Factors in
  Computing Systems, {CHI} 2008, 2008, Florence, Italy, April 5-10, 2008},
  pages 453--456. {ACM}.

\bibitem[{Lample et~al.(2016)Lample, Ballesteros, Subramanian, Kawakami, and
  Dyer}]{lample-etal-2016-neural}
Guillaume Lample, Miguel Ballesteros, Sandeep Subramanian, Kazuya Kawakami, and
  Chris Dyer. 2016.
\newblock \href {https://doi.org/10.18653/v1/N16-1030} {Neural architectures
  for named entity recognition}.
\newblock In \emph{Proceedings of NAACL-HLT}, pages 260--270.

\bibitem[{Lease(2011)}]{lease2011on}
Matthew Lease. 2011.
\newblock \href {http://www.aaai.org/ocs/index.php/WS/AAAIW11/paper/view/3906}
  {On quality control and machine learning in crowdsourcing}.
\newblock In \emph{Human Computation}, volume {WS-11-11} of \emph{{AAAI}
  Workshops}. {AAAI}.

\bibitem[{Li et~al.(2020)Li, Takamura, and Ananiadou}]{li-etal-2020-neural}
Maolin Li, Hiroya Takamura, and Sophia Ananiadou. 2020.
\newblock \href {https://doi.org/10.18653/v1/2020.coling-main.507} {A neural
  model for aggregating coreference annotation in crowdsourcing}.
\newblock In \emph{Proceedings of COLING}, pages 5760--5773.

\bibitem[{Liu et~al.(2017)Liu, Jiang, Liu, Wang, Zhu, and
  Liu}]{liu2017improving}
Mengchen Liu, Liu Jiang, Junlin Liu, Xiting Wang, Jun Zhu, and Shixia Liu.
  2017.
\newblock \href {https://doi.org/10.24963/ijcai.2017/324} {Improving
  learning-from-crowds through expert validation}.
\newblock In \emph{Proceedings of IJCAI}, pages 2329--2336.

\bibitem[{Manning and Schutze(1999)}]{manning1999foundations}
Christopher Manning and Hinrich Schutze. 1999.
\newblock \emph{Foundations of statistical natural language processing}.
\newblock MIT press.

\bibitem[{Mansour et~al.(2009)Mansour, Mohri, and
  Rostamizadeh}]{mansour2009domain}
Yishay Mansour, Mehryar Mohri, and Afshin Rostamizadeh. 2009.
\newblock \href
  {https://proceedings.neurips.cc/paper/2008/file/0e65972dce68dad4d52d063967f0a705-Paper.pdf}
  {Domain adaptation with multiple sources}.
\newblock In \emph{Advances in Neural Information Processing Systems},
  volume~21, pages 1041--1048.

\bibitem[{Mayhew et~al.(2020)Mayhew, Gupta, and Roth}]{stephen2020robust}
Stephen Mayhew, Nitish Gupta, and Dan Roth. 2020.
\newblock \href {https://aaai.org/ojs/index.php/AAAI/article/view/6368} {Robust
  named entity recognition with truecasing pretraining}.
\newblock In \emph{{AAAI} 2020}, pages 8480--8487.

\bibitem[{Nguyen et~al.(2017)Nguyen, Wallace, Li, Nenkova, and
  Lease}]{nguyen-etal-2017-aggregating}
An~Thanh Nguyen, Byron Wallace, Junyi~Jessy Li, Ani Nenkova, and Matthew Lease.
  2017.
\newblock \href {https://doi.org/10.18653/v1/P17-1028} {Aggregating and
  predicting sequence labels from crowd annotations}.
\newblock In \emph{Proceedings of ACL}, pages 299--309.

\bibitem[{Nye et~al.(2018)Nye, Li, Patel, Yang, Marshall, Nenkova, and
  Wallace}]{nye-etal-2018-corpus}
Benjamin Nye, Junyi~Jessy Li, Roma Patel, Yinfei Yang, Iain Marshall, Ani
  Nenkova, and Byron Wallace. 2018.
\newblock \href {https://doi.org/10.18653/v1/P18-1019} {A corpus with
  multi-level annotations of patients, interventions and outcomes to support
  language processing for medical literature}.
\newblock In \emph{Proceedings of the ACL}, pages 197--207.

\bibitem[{Platanios et~al.(2018)Platanios, Sachan, Neubig, and
  Mitchell}]{platanios-etal-2018-contextual}
Emmanouil~Antonios Platanios, Mrinmaya Sachan, Graham Neubig, and Tom Mitchell.
  2018.
\newblock \href {https://doi.org/10.18653/v1/D18-1039} {Contextual parameter
  generation for universal neural machine translation}.
\newblock In \emph{Proceedings of EMNLP}.

\bibitem[{Ramponi and Plank(2020)}]{ramponi-plank-2020-neural}
Alan Ramponi and Barbara Plank. 2020.
\newblock \href {https://doi.org/10.18653/v1/2020.coling-main.603} {Neural
  unsupervised domain adaptation in {NLP}{---}{A} survey}.
\newblock In \emph{Proceedings of the COLING}, pages 6838--6855.

\bibitem[{Raykar and Yu(2012{\natexlab{a}})}]{raykar2012eli}
Vikas~C. Raykar and Shipeng Yu. 2012{\natexlab{a}}.
\newblock \href {http://dl.acm.org/citation.cfm?id=2188401} {Eliminating
  spammers and ranking annotators for crowdsourced labeling tasks}.
\newblock \emph{J. Mach. Learn. Res.}, 13:491--518.

\bibitem[{Raykar and Yu(2012{\natexlab{b}})}]{vikas2012eliminating}
Vikas~C. Raykar and Shipeng Yu. 2012{\natexlab{b}}.
\newblock \href {http://dl.acm.org/citation.cfm?id=2188401} {Eliminating
  spammers and ranking annotators for crowdsourced labeling tasks}.
\newblock \emph{J. Mach. Learn. Res.}, 13:491--518.

\bibitem[{Raykar et~al.(2010)Raykar, Yu, Zhao, Valadez, Florin, Bogoni, and
  Moy}]{raykar2010learning}
Vikas~C. Raykar, Shipeng Yu, Linda~H. Zhao, Gerardo~Hermosillo Valadez, Charles
  Florin, Luca Bogoni, and Linda Moy. 2010.
\newblock \href {http://portal.acm.org/citation.cfm?id=1859894} {Learning from
  crowds}.
\newblock \emph{J. Mach. Learn. Res.}, 11:1297--1322.

\bibitem[{Rodrigues and Pereira(2018)}]{rodrigues2018deep}
Filipe Rodrigues and Francisco~C. Pereira. 2018.
\newblock \href
  {https://www.aaai.org/ocs/index.php/AAAI/AAAI18/paper/view/16102} {Deep
  learning from crowds}.
\newblock In \emph{Proceedings of the AAAI}.

\bibitem[{Rodrigues et~al.(2014)Rodrigues, Pereira, and
  Ribeiro}]{rodrigues2014sequence}
Filipe Rodrigues, Francisco~C. Pereira, and Bernardete Ribeiro. 2014.
\newblock \href {https://doi.org/10.1007/s10994-013-5411-2} {Sequence labeling
  with multiple annotators}.
\newblock \emph{Mach. Learn.}, 95(2):165--181.

\bibitem[{Sener et~al.(2016)Sener, Song, Saxena, and
  Savarese}]{sener2016learning}
Ozan Sener, Hyun~Oh Song, Ashutosh Saxena, and Silvio Savarese. 2016.
\newblock \href
  {https://proceedings.neurips.cc/paper/2016/hash/b59c67bf196a4758191e42f76670ceba-Abstract.html}
  {Learning transferrable representations for unsupervised domain adaptation}.
\newblock In \emph{Advances in Neural Information Processing Systems}.

\bibitem[{Sheng et~al.(2008)Sheng, Provost, and Ipeirotis}]{sheng2008get}
Victor~S. Sheng, Foster~J. Provost, and Panagiotis~G. Ipeirotis. 2008.
\newblock \href {https://doi.org/10.1145/1401890.1401965} {Get another label?
  improving data quality and data mining using multiple, noisy labelers}.
\newblock In \emph{Proceedings of the KDD}, pages 614--622.

\bibitem[{Sheng and Zhang(2019)}]{Sheng_Zhang_2019}
Victor~S. Sheng and Jing Zhang. 2019.
\newblock \href {https://doi.org/10.1609/aaai.v33i01.33019837} {Machine
  learning with crowdsourcing: A brief summary of the past research and future
  directions}.
\newblock \emph{Proceedings of the AAAI}, 33(01):9837--9843.

\bibitem[{Simpson and Gurevych(2019)}]{simpson-gurevych-2019-bayesian}
Edwin Simpson and Iryna Gurevych. 2019.
\newblock \href {https://doi.org/10.18653/v1/D19-1101} {A {B}ayesian approach
  for sequence tagging with crowds}.
\newblock In \emph{Proceedings of the EMNLP-IJCNLP}.

\bibitem[{Snow et~al.(2008)Snow, O{'}Connor, Jurafsky, and
  Ng}]{snow-etal-2008-cheap}
Rion Snow, Brendan O{'}Connor, Daniel Jurafsky, and Andrew Ng. 2008.
\newblock \href {https://www.aclweb.org/anthology/D08-1027} {Cheap and fast
  {--} but is it good? evaluating non-expert annotations for natural language
  tasks}.
\newblock In \emph{Proceedings of the EMNLP}.

\bibitem[{Tang and Lease(2011)}]{tang2011semi}
Wei Tang and Matthew Lease. 2011.
\newblock Semi-supervised consensus labeling for crowdsourcing.
\newblock In \emph{SIGIR 2011 workshop on crowdsourcing for information
  retrieval (CIR)}, pages 1--6.

\bibitem[{Tjong Kim~Sang and
  De~Meulder(2003)}]{tjong-kim-sang-de-meulder-2003-introduction}
Erik~F. Tjong Kim~Sang and Fien De~Meulder. 2003.
\newblock \href {https://www.aclweb.org/anthology/W03-0419} {Introduction to
  the {C}o{NLL}-2003 shared task: Language-independent named entity
  recognition}.
\newblock In \emph{Proceedings of the CoNLL at {HLT}-{NAACL} 2003}.

\bibitem[{Trautmann et~al.(2020)Trautmann, Daxenberger, Stab, Sch{\"{u}}tze,
  and Gurevych}]{dietrich2020fine}
Dietrich Trautmann, Johannes Daxenberger, Christian Stab, Hinrich
  Sch{\"{u}}tze, and Iryna Gurevych. 2020.
\newblock \href {https://aaai.org/ojs/index.php/AAAI/article/view/6438}
  {Fine-grained argument unit recognition and classification}.
\newblock In \emph{{AAAI} 2020}, pages 9048--9056. {AAAI} Press.

\bibitem[{{\"U}st{\"u}n et~al.(2020){\"U}st{\"u}n, Bisazza, Bouma, and van
  Noord}]{ustun-etal-2020-udapter}
Ahmet {\"U}st{\"u}n, Arianna Bisazza, Gosse Bouma, and Gertjan van Noord. 2020.
\newblock \href {https://doi.org/10.18653/v1/2020.emnlp-main.180} {{UD}apter:
  Language adaptation for truly {U}niversal {D}ependency parsing}.
\newblock In \emph{Proceedings of the EMNLP}, pages 2302--2315.

\bibitem[{Wang et~al.(2020)Wang, Kulkarni, and
  Preotiuc-Pietro}]{wang-etal-2020-multi-domain}
Jing Wang, Mayank Kulkarni, and Daniel Preotiuc-Pietro. 2020.
\newblock \href {https://doi.org/10.18653/v1/2020.acl-main.750} {Multi-domain
  named entity recognition with genre-aware and agnostic inference}.
\newblock In \emph{Proceedings of the ACL}, pages 8476--8488.

\bibitem[{Wang and Zhou(2016)}]{wang2016cost}
Lu~Wang and Zhi-Hua Zhou. 2016.
\newblock Cost-saving effect of crowdsourcing learning.
\newblock In \emph{Proceedings of IJCAI}, IJCAI'16, pages 2111--2117.

\bibitem[{Wolf et~al.(2020)Wolf, Debut, Sanh, Chaumond, Delangue, Moi, Cistac,
  Rault, Louf, Funtowicz, Davison, Shleifer, von Platen, Ma, Jernite, Plu, Xu,
  Le~Scao, Gugger, Drame, Lhoest, and Rush}]{wolf-etal-2020-transformers}
Thomas Wolf, Lysandre Debut, Victor Sanh, Julien Chaumond, Clement Delangue,
  Anthony Moi, Pierric Cistac, Tim Rault, Remi Louf, Morgan Funtowicz, Joe
  Davison, Sam Shleifer, Patrick von Platen, Clara Ma, Yacine Jernite, Julien
  Plu, Canwen Xu, Teven Le~Scao, Sylvain Gugger, Mariama Drame, Quentin Lhoest,
  and Alexander Rush. 2020.
\newblock \href {https://doi.org/10.18653/v1/2020.emnlp-demos.6} {Transformers:
  State-of-the-art natural language processing}.
\newblock In \emph{Proceedings of the EMNLP: System Demonstrations}, pages
  38--45.

\bibitem[{Yadav and Bethard(2018)}]{yadav-bethard-2018-survey}
Vikas Yadav and Steven Bethard. 2018.
\newblock \href {https://www.aclweb.org/anthology/C18-1182} {A survey on recent
  advances in named entity recognition from deep learning models}.
\newblock In \emph{Proceedings of the COLING}.

\bibitem[{Yang et~al.(2018)Yang, Zhang, Chen, Zhang, Wang, and
  Zhang}]{yang2018adversarial}
YaoSheng Yang, Meishan Zhang, Wenliang Chen, Wei Zhang, Haofen Wang, and Min
  Zhang. 2018.
\newblock \href
  {https://www.aaai.org/ocs/index.php/AAAI/AAAI18/paper/view/16427}
  {Adversarial learning for chinese {NER} from crowd annotations}.
\newblock In \emph{Proceedings of the AAAI}.

\bibitem[{Yu et~al.(2015)Yu, Elkaref, and Bohnet}]{yu-etal-2015-domain}
Juntao Yu, Mohab Elkaref, and Bernd Bohnet. 2015.
\newblock \href {https://doi.org/10.18653/v1/W15-2201} {Domain adaptation for
  dependency parsing via self-training}.
\newblock In \emph{Proceedings of the 14th International Conference on Parsing
  Technologies}, pages 1--10.

\bibitem[{Zaidan and
  Callison-Burch(2011)}]{zaidan-callison-burch-2011-crowdsourcing}
Omar~F. Zaidan and Chris Callison-Burch. 2011.
\newblock \href {https://www.aclweb.org/anthology/P11-1122} {Crowdsourcing
  translation: Professional quality from non-professionals}.
\newblock In \emph{Proceedings of the ACL-HLT}, pages 1220--1229.

\bibitem[{Zhang et~al.(2016)Zhang, Wu, and Sheng}]{zhang2016learning}
Jing Zhang, Xindong Wu, and Victor~S. Sheng. 2016.
\newblock \href {https://doi.org/10.1007/s10462-016-9491-9} {Learning from
  crowdsourced labeled data: a survey}.
\newblock \emph{Artif. Intell. Rev.}, 46(4):543--576.

\bibitem[{Zhao et~al.(2019)Zhao, des Combes, Zhang, and Gordon}]{zhao2019on}
Han Zhao, Remi~Tachet des Combes, Kun Zhang, and Geoffrey~J. Gordon. 2019.
\newblock \href {http://proceedings.mlr.press/v97/zhao19a.html} {On learning
  invariant representations for domain adaptation}.
\newblock In \emph{Proceedings of the ICML}, pages 7523--7532. {PMLR}.

\end{thebibliography}

\appendix

\begin{figure} \centering
\includegraphics{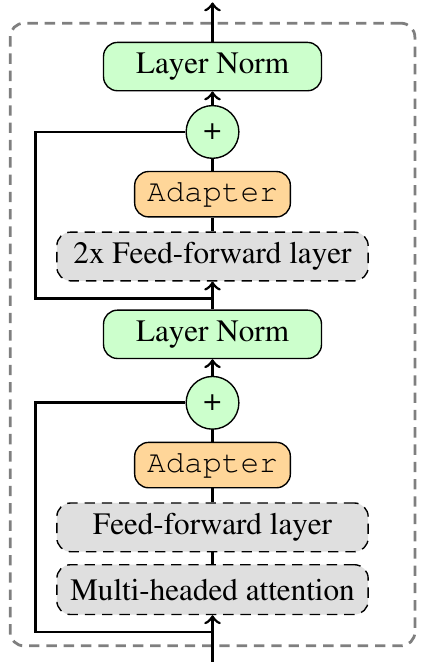}
\caption{Transformer integrated with Adapters inside.}
\label{figure:layer}
\end{figure}

\section{Transformer with Adapters}
\label{appendix:transformer}
In our $\text{Adapter}\circ\text{BERT}$ word representation,
we insert two adapter modules for each transformer layer inside BERT.
Figure \ref{figure:layer} shows the detailed network structure of transformer with adapters.
More specifically, the forward operation of an adapter layer is computed as follows:
\begin{equation} \label{adapter}
\begin{split}
    &  \bm{h}_{\text{mid}} = \mathrm{GELU}(\bm{W}^{\text{ap}}_1\bm{h}_{\text{in}} + \bm{b}^{\text{ap}}_1) \\
    &  \bm{h}_{\text{out}} = \bm{W}^{\text{ap}}_2\bm{h}_{\text{mid}} + \bm{b}^{\text{ap}}_2 + \bm{h}_{\text{in}},
\end{split}
\end{equation}
where $\bm{W}^{\text{ap}}_1$, $\bm{W}^{\text{ap}}_2$, $\bm{b}^{\text{ap}}_1$ and $\bm{b}^{\text{ap}}_2$ are adapter parameters,
and the dimension size of $\bm{h}_{\text{mid}}$ is usually smaller than that of the corresponding transformer.

Here we also give a supplement to illustrate the pack operation from all adapter parameters into a single vector $\bm{V}$:
\begin{equation} \label{ada-param}
    \bm{V} = \bigoplus_{\text{Adapters}}\{\bm{W}^{\text{ap}}_1 \oplus \bm{W}^{\text{ap}}_2 \oplus \bm{b}^{\text{ap}}_1 \oplus \bm{b}^{\text{ap}}_2\},
\end{equation}
where first all parameters of a single adapter are reshaped and concatenated and then a further concatenation is performed over all adapters.

\setlength{\tabcolsep}{6.0pt}
\begin{table} \centering \small
\begin{tabular}{c|ccc|c}
\hline
\multirow{2}{*}{Model} & \multirow{2}{*}{ALL} & \multirow{2}{*}{MV} & \multirow{2}{*}{Gold} & \multirow{2}{*}{\begin{tabular}[c]{@{}c@{}}Trainable\\ Params Size \end{tabular}}  \\
 &   &   &   &   \\
\hline \hline
FineTuning & 74.12 & 74.96 & 89.32 & \bf 108M \\
\hline \hline
\multicolumn{5}{c}{BERT with Adapter Inside} \\
\hline \hline
2 layers    & 71.83 & 73.81 & 89.20  & 4.55M \\ 
4 layers    & 73.16 & 73.30 & 89.26  & 5.34M \\ 
6 layers    & 73.74 & 74.81 & 89.33  & 6.14M \\ 
8 layers    & 74.24 & \bf 75.31 & 89.13  & 6.94M \\  
10 layers   & \bf 74.56 & 75.01 & 89.21  & 7.73M \\  
All layers   & 74.36 & 75.28 & \bf 89.52  & 8.53M \\  
\hline
\end{tabular}
\caption{The comparisons between BERT fine-tuning and $\text{Adapter}\circ\text{BERT}$ based on the standard NER without annotator as input.}
\label{table:parameter}
\end{table}

\renewcommand\arraystretch{1.0}
\definecolor{a1}{HTML}{0F0F00}
\definecolor{a2}{HTML}{E56709}
\begin{table*} \centering \small
\begin{tabular}{c|p{12cm}} 
\toprule
\bf Model & \multicolumn{1}{c}{\bf Text and Entities} \\  
\hline\hline
\multicolumn{2}{c}{\bf Unsupervised} \\
\hline\hline
\multirow{2}{*}{MV}   &
$\underline{\text{Pace}}$, a junior, helped \textcolor{red}{$[${\bf Ohio State}}$\textcolor{red}{]}_{\underline{LOC}}$
to a 10-1 record and a berth in the \underline{Rose Bowl}
against \underline{$[$Arizona$]_{ORG}$ State}. \\
\multirow{2}{*}{LC-cat}    &
$\underline{\text{Pace}}$, a junior, helped \textcolor{red}{$[${\bf Ohio State}$]_{ORG}$}
to a 10-1 record and a berth in the  \textcolor{a1}{$[${\bf Rose Bowl}$]_{MISC}$}
against \underline{$[$Arizona$]_{ORG}$ State}. \\
\multirow{2}{*}{This Work} &
$\underline{\text{Pace}}$, a junior, helped \textcolor{red}{$[${\bf Ohio State}$]_{ORG}$}
to a 10-1 record and a berth in the  \textcolor{a1}{$[${\bf Rose Bowl}$]_{MISC}$}
against \textcolor{a2}{$[${\bf Arizona State}$]_{ORG}$}. \\
\hline\hline
\multicolumn{2}{c}{\bf Supervised (25\%)} \\
\hline\hline
\multirow{2}{*}{MV}   &
$\underline{\text{Pace}}$, a junior, helped \textcolor{red}{$[${\bf Ohio State}}$\textcolor{red}{]}_{\underline{LOC}}$
to a 10-1 record and a berth in the  \textcolor{a1}{$[${\bf Rose Bowl}$]_{MISC}$}
against \textcolor{a2}{$[${\bf Arizona State}}$\textcolor{a2}{]}_{\underline{LOC}}$. \\
\multirow{2}{*}{Gold}  &
\textcolor{blue}{$[${\bf Pace}$]_{PER}$}, a junior, helped \textcolor{red}{$[${\bf Ohio State}$]_{ORG}$}
to a 10-1 record and a berth in the  \textcolor{a1}{$[${\bf Rose Bowl}$]_{MISC}$}
against \underline{$[$Arizona$]_{ORG}$ State}. \\
\multirow{2}{*}{LC-cat}    &
$\underline{\text{Pace}}$, a junior, helped \textcolor{red}{$[${\bf Ohio State}$]_{ORG}$}
to a 10-1 record and a berth in the  \textcolor{a1}{$[${\bf Rose Bowl}$]_{MISC}$}
against \textcolor{a2}{$[${\bf Arizona State}}$\textcolor{a2}{]}_{\underline{LOC}}$. \\
\multirow{2}{*}{This Work} &
\textcolor{blue}{$[${\bf Pace}$]_{PER}$}, a junior, helped \textcolor{red}{$[${\bf Ohio State}$]_{ORG}$}
to a 10-1 record and a berth in the  \textcolor{a1}{$[${\bf Rose Bowl}$]_{MISC}$}
against \textcolor{a2}{$[${\bf Arizona State}$]_{ORG}$}. \\
\hline\hline
\multirow{2}{*}{Ground-truth} &
\textcolor{blue}{$[${\bf Pace}$]_{PER}$}, a junior, helped \textcolor{red}{$[${\bf Ohio State}$]_{ORG}$}
to a 10-1 record and a berth in the  \textcolor{a1}{$[${\bf Rose Bowl}$]_{MISC}$}
against \textcolor{a2}{$[${\bf Arizona State}$]_{ORG}$}. \\
\bottomrule
\end{tabular}
\caption{A case study, where the text with underlines indicates errors.}
\label{tab:case:study}
\end{table*}

\section{Hyper-parameters}\label{appendix:hyper-param}
We choose the BERT-base-cased\footnote{https://github.com/google-research/bert},
which is for English language and consists of 12-layer transformers with the hidden size 768 for all layers.
We load the BERT weight and implement the adapter injection based on the transformers \cite{wolf-etal-2020-transformers} library.
The sizes of the adapter middle hidden states are set to 128 constantly.
The annotator embedding size is 8 to fit the model in one RTX-2080TI GPU of 11GB memory.
The BiLSTM hidden size is set to 400.
For all models, we inject adapters or switchers in all 12 layers of BERT.
All experiments are run on the single GPU at an 8-GPU server with a 14 core CPU and 128GB memory.

We exploit the stochastic gradient-based online learning, with a batch size of 64, to optimize model parameters.
We apply the time-step dropout, which randomly sets several representations in the sequence to zeros with a probability of $0.2$,
on the word representations to avoid overfitting.
We use the Adam algorithm to update the parameters with a constant learning rate $1 \times 10^{-3}$,
and apply the gradient clipping by a maximum value of $5.0$ to avoid gradient explosion.

\section{The Advantage of $\text{Adapter}\circ\text{BERT}$ }
Our models are all based on $\text{Adapter}\circ\text{BERT}$ as the basic representations,
which is different from the widely-adopted BERT fine-tuning architecture.
Here we compare the two strategies in detail.
The results are shown in Table \ref{table:parameter},
where for $\text{Adapter}\circ\text{BERT}$ we consider
gradually increasing the number of transformer layers
(covering the last $n$ layers) inside the BERT.
As shown,
it is apparently that $\text{Adapter}\circ\text{BERT}$ is much more parameter efficient,
and when all layers are exploited, the model can be even better than BERT fine-tuning.
Thus it is more desirable to use $\text{Adapter}\circ\text{BERT}$ covering all BERT transformers inside.

\section{Case Study}
Here we also offer a case study to understand the performance in unsupervised and supervised crowdsourcing learning,
as well as the different crowdsourcing models.
We exploit one complex example in Table \ref{tab:case:study} which involves different outputs for various models.
As shown, we can see that supervised models are able to recall the ambiguous entity (i.e., Pace, a single word with multiple senses) correctly,
while unsupervised models fail,
which may be due to the inconsistencies of the crowdsourced annotations. 
By comparing our model with other baselines,
we can show that our representation learning model can capture the global text input understanding consistently,
e.g., being able to connect Ohio State and Arizona State together.

\end{document}